\documentclass[letterpaper, 10 pt, conference]{ieeeconf}  % Comment this line out if you need a4paper

\IEEEoverridecommandlockouts                              % This command is only needed if 
\usepackage{graphicx}
\usepackage{subfigure}
\usepackage[hidelinks]{hyperref}
\usepackage{amsmath}
\usepackage[dvipsnames,table]{xcolor}
\usepackage{multirow}
\usepackage{tabularx}
\usepackage{threeparttable}
\usepackage[capitalize]{cleveref}
\usepackage{booktabs}

\usepackage{adjustbox}

\usepackage{tikz}
\usetikzlibrary{positioning,calc,matrix,fit,arrows.meta,shapes}

\pgfdeclarelayer{background}
\pgfdeclarelayer{foreground}
\pgfsetlayers{background,main,foreground}

\usepackage{pgfplots}
\pgfplotsset{compat=1.18}

\definecolor{anasgreen}{HTML}{b4e5a2}

\title{\LARGE \bf
Learning Embeddings with Centroid Triplet Loss\\
for Object Identification in Robotic Grasping
}
\author{Anas Gouda$^{1,3}$, Max Schwarz$^{2,3}$, Christopher Reining$^{1}$, Sven Behnke$^{2,3}$, and Alice Kirchheim$^{1,3,4}$% <-this % stops a space
\thanks{\hspace{-.1em}$^{1}$TU Dortmund,
        {\tt\small anas.gouda@tu-dortmund.de}}%
\thanks{\hspace{-.1em}$^{2}$Autonomous Intelligent Systems - Computer Science VI \&\linebreak\hspace{-.1em}\phantom{$^{2}$} Center for Robotics, University of Bonn, Germany}%
\thanks{\hspace{-.1em}$^{3}$Lamarr Institute for Machine Learning and Artificial Intelligence}
\thanks{\hspace{-.1em}$^{4}$Fraunhofer IML}
}%

\begin{document}

\maketitle
\thispagestyle{empty}
\pagestyle{empty}
\setcounter{footnote}{4}

%%%%%%%%%%%%%%%%%%%%%%%%%%%%%%%%%%%%%%%%%%%%%%%%%%%%%%%%%%%%%%%%%%%%%%%%%%%%%%%%
\begin{abstract}

% Introduction
% Foundation models for object segmentation are trending
Foundation models are a strong trend in deep learning and computer vision. These models serve as a base for applications as they require minor or no further fine-tuning by developers to integrate into their applications.
%We follow this trend with a focus on object segmentation in robotics grasping.
%% Problem Statement
% why do we mention foundation models?
Foundation models for zero-shot object segmentation such as Segment Anything (SAM) output segmentation masks from images without any further object information. When they are followed in a pipeline by an object identification model, they can perform object detection without training.
%Using only few images of an object such a pipeline would detect and segment the object.
%Here we focus on a training a model for object identification for use in such pipelines.
Here, we focus on training such an object identification model.
%% Methodology
% object identification models should generalize well and be flexible
% big datasets were missing
% CTL is the solution
%Large-scale datasets such as ArmBench and MegaPose allow for training such models for object identification.
A crucial practical aspect for an object identification model is to be flexible in input size (number of input images). As object identification is an image retrieval problem, a suitable method should handle multi-query multi-gallery situations without constraining the number of input images (e.g. by having fixed-size aggregation layers).
The key solution to train such a model is the centroid triplet loss (CTL), which aggregates image features to their centroids. CTL yields high accuracy, avoids misleading training signals and keeps the model input size flexible.
%% Key Findings, Implications and Contributions
In our experiments, we establish a new state of the art on the ArmBench object identification task, which shows general applicability of our model.
%On the ArmBench test set we achieve the highest score of 98.7\% which proves that our model can be used directly out of the box in many applications.
We furthermore demonstrate an integrated unseen object detection pipeline on the challenging HOPE dataset, which requires fine-grained detection. There, our pipeline matches and surpasses related methods which have been trained on dataset-specific data.
%On the challenging fine-grained HOPE dataset (which represent the hard limits for our model) we score 0.45 AP50. This makes our model which is not fine-tuned on the HOPE object surpass trained methods such as Mask R-CNN that is trained on the HOPE dataset objects.
Code and pretrained models are available.\footnote{%
\scriptsize\url{https://github.com/AnasIbrahim/ctl_classification}%
}
%To use our model directly in an application refer to our DoUnseen package%
%\footnote{\tiny
%\url{https://github.com/AnasIbrahim/image_agnostic_segmentation}
%\label{footnote:github_dounseen}
%}.
%The training and evaluation scripts are also available%
%\footnote{\tiny
%\url{https://github.com/AnasIbrahim/ctl_classification}
%\label{footnote:github_ctl}
%}.

\end{abstract}

%%%%%%%%%%%%%%%%%%%%%%%%%%%%%%%%%%%%%%%%%%%%%%%%%%%%%%%%%%%%%%%%%%%%%%%%%%%%%%%%
\section{Introduction}

\begin{figure}
 \input{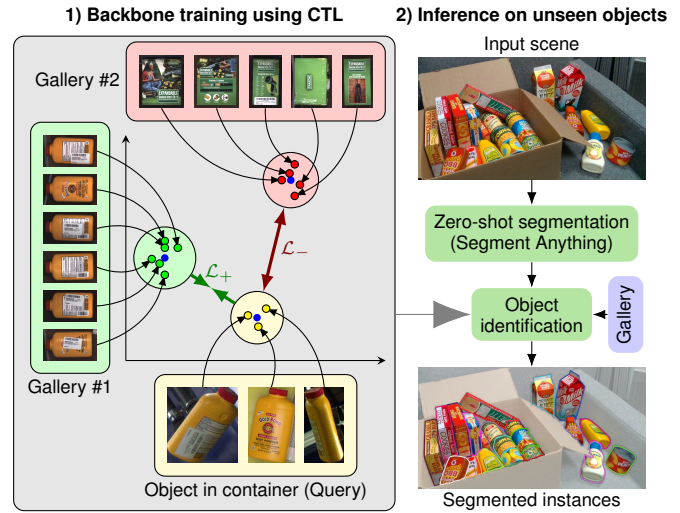}
 \vspace{-3ex}
 \caption{Overview of our method. We train an object identification backbone on the large-scale ARMBench ID dataset
 using the centroid triplet loss (CTL~\cite{unreasonable_ctl}), such that the backbone learns to associate query images of objects in cluttered containers
 to matching gallery images.
 Crucially, the CTL loss operates on centroids in feature space, allowing the aggregation
 of an arbitrary number of input images.
 The trained backbone can then be used for identification of unseen objects that were segmented using a generic
 object segmentation method such as Segment Anything (SAM), given corresponding gallery images.
 }
 \label{fig:ctl}
\end{figure}

% 1) Motivation / Relevance:
% Many logistics entities and retailers in industry capture image of their products. Walmart for example offers an API service to acquire images of any of the products by their product ID. Companies such as Arvato capture images of all incoming new unique products to guide workers in their picking stations \cite{}. These images represent the gallery images. Having such images empowers to enhance or develop some applications such as:
% \begin{itemize}
%     \item Verify grasped objects:
%     A mistake that happens often in off-the-shelf software for bin picking is grasping dummy objects. in a realistic case, a bin can contain leftovers from product packages or empty cardboard boxes. Zero-shot segmentation would blindly segment them. A generic object-matching model with images of the product can verify if the grasped objects are correct.
%     \item Mixed bin picking:
%     This is a well-known application that came up in the Amazon robotics challenge \cite{}. Where many objects of different types are stored together in one bin.
%     \item inventory counting and defect detection:
%     The process could be automated by associating the object in the database with the real content of the bin.
%     \item Mixed material flow:
%     from an economical perspective and logistics process optimization, it can be cheaper or faster when multiple orders are picked together from a bin, and the flow gets separated later.
% \end{itemize}

% General Motivation
Object perception is a crucial prerequisite for many logistics applications,  
such as mixed bin picking, which received attention in the Amazon Robotics Challenge~\cite{morrison2018cartman, SchwarzLGKPSB:ICRA18}.
Product verification is another application where object identification is required to eliminate mistakes, e.g., robots mistakenly picking dummy objects like packaging material. 
Further use cases include multi-order picking and handling of returned goods.

The number of unique objects handled along supply chains reaches millions, posing a significant challenge to object perception systems.
Mainly due to the limitations of current object perception methods, many of the above use cases are still not automated, but performed by human operators.

% Problem Statement and the potential how to solve it
While deep-learning based methods have potential to enable automation of these applications, training data scarcity has prevented their breakthrough so far.
Recent releases of public large-scale datasets such as ARMBench~\cite{armbench} and MegaPose~\cite{megapose} are potential game changers. 
In addition, images and data from warehouses themselves become available to researchers,
as many retailers capture and offer images of all of their products. 
Some, such as Walmart, even provide API to retrieve these images for general use. 
These images could serve as reference data to identify objects in warehouse scenes.

% State of the Art --> specify research gap
Traditional methods for instance segmentation (such as Mask R-CNN~\cite{maskrcnn} and its derivatives) assume an object set that is fixed at training time.
This is a serious limitation as any change in the set of objects requires extensive retraining. Fine-tuning of segmentation models can mitigate this issue to some extent, but it requires careful attention and is still expensive.
To address object set variability, methods for zero-shot and few-shot object identification are a suitable approach, as they do not require any adaptation.
%All the previously mentioned applications of robotic grasping share a common need for two different backbone modules --- zero-shot segmentation and object identification.
Recently, there have been big advancements in zero-shot object segmentation~\cite{segment_anything,fast_sam}.
Other advancements include category-agnostic template matching~\cite{template_matching_Hu} and object identification based on multimodal large language models~\cite{robollm}.
% The methods' shortcoming is their lack of flexibility in the input size.
There are several shortcomings in these methods, though. Recent template matching methods as~\cite{template_matching_Hu} handle only a single type of object; this is due to template images being fed early in the model layers. To detect multiple different object types, the methods need to be run repeatedly. This leads to long execution times that are unpractical for some applications. Furthermore, template matching methods are hard to enhance as the segmentation and object identification/matching are developed as a single black box. Fine-tuning a model to eliminate specific drawbacks might not be possible. Methods for object identification such as RoboLLM~\cite{robollm} tailor their query-input size to the application. If the number of query images changes, the model needs to be retrained. This retraining might not even be possible for applications where query images are collected automatically, which is a common practice in many of off-the-shelf-software. RoboLLM can also handle only one query object at a time.

% Goal
In this paper, we assume the availability of a generic zero-shot segmentation method and focus on the object identification task, i.e. determining which class a segment belongs to. 
Our objective is to develop a method for object identification that is flexible in the number of input images and scalable in the number of objects.
\cref{fig:ctl} illustrates our approach for object identification in the context of robotic grasping.
We train a backbone network that maps object images to embeddings in an abstract feature space using a centroid triplet loss (CTL)~\cite{unreasonable_ctl}. 
In this feature space, our approach matches pre-captured object images (gallery images) to query images, which are generated by a zero-shot segmentation model or by application-specific segmentation models.
Our method allows for processing any number of gallery and query objects, both described by an arbitrary number of images.

% Method and Contribution
%Our model outperforms previous work on the ARMBench object identification test set. Combined with Segment-Anything (SAM)~\cite{segment_anything}, our pipeline outperforms trained methods on the HOPE dataset~\cite{hope_dataset}.
\noindent Our contributions include:
\begin{enumerate}
    \item an approach for training object identification backbones with the centroid triplet loss on large-scale datasets,
    \item evaluation of the backbones on ARMBench, where we establish a new state of the art,
    \item an integrated architecture for unseen object instance segmentation with said backbone, and
    \item evaluation and ablation of the entire pipeline on the HOPE dataset, where we obtain comparable performance to a method trained with object information.
    %\item Our model serves as a backbone to enhance the performance of pipelines for 2D segmentation of unseen objects such as \cite{cnos} \cite{dounseen} or pipelines for 6D localization of unseen objects such as \cite{sam6d}.
    %\item Also our model can be used standalone for several of the pre-mentioned applications.
\end{enumerate}

% With this object identification model we hit two birds with one stone.
% First, we handle the aforementioned application without having to even fine-tune the model. 
% Second, we offer this model as the object identification/matching backbone for unseen 6D pose estimation pipelines like SAM-6D.
% Our contribution is an experimental evaluation how to use the centroid triplet loss to train a model for flexible and scalable object identification. 
% We use the ARMBench object identification dataset \cite{armbench} for our training.
% We train multiple architectures with the centroid triplet loss (CTL) (see Fig.~\ref{fig:ctl}).
% %It provides a way to train the model with the ARMBench dataset as shown in figure \ref{fig:ctl}.
% All object query and gallery images are used at once for each batch. That way object all object data are fed during training to the network at once and, most importantly, the model stays flexible during testing without having any fixed size MLP layers.

% Outline of this paper
In Section~\ref{sec:related}, we discuss related work in detail.
%on the bigger picture of  how our object identification model serves pipelines for unseen object detection.
In Section~\ref{sec:method}, we provide details on the centroid triplet loss and the training process for the ARMBench dataset. 
We report the evaluation results in Section~\ref{sec:results}.
\section{Related Work}
\label{sec:related}

\begin{figure}
    \vspace{0.2cm}
    \centering
    \includegraphics[width=0.48\textwidth, trim={250 295 240 230}, clip]{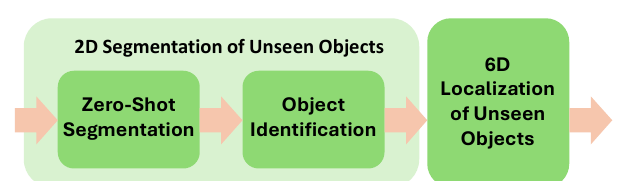}
    \caption{Typical stages for handling unseen objects. Zero-shot segmentation and object identification/matching represent by themselves a pipeline for 2D segmentation of unseen objects. Another stage is then optionally added to perform 6D pose estimation of unseen objects.
    %This excludes render and compare methods for 6D Pose of unseen objects.
    }
    \label{fig:unseen_6d_pipelines}
\end{figure}

Pipelines working with unseen objects typically include multiple stages with separate deep neural networks.
As shown in \cref{fig:unseen_6d_pipelines}, typically these stages are zero-shot object segmentation, object identification or matching, and 6D localization of unseen objects. The first and second stage combined represent a module by themselves for 2D segmentation of unseen objects.
Large-scale datasets with high variation are also a pillar for training these DNNs.
This section provides an overview of current research in each stage and elaborates on its limitations.

% zero-shot object segmentation (keep it short)
%1. describe what researchers are already doing. Use the term above specifically (or explain synonyms)
%2. highlight the interaction with your contribution. do you use it?
%3. foreshadowing: what would be the next step here (that will be taken in this contribution). or does your contribution build on it?

% object classification
%1. describe what researchers are already doing. Use the term above specifically (or explain synonyms)
%In Object Classification, sometimes also referred to as xxx or xxx, huge progress was made with regards to ...
%2. highlight the limitation / shortcoming
%3. foreshadowing: what would be the next step here (that will be taken in this contribution). or does your contribution build on it?

% 6D Pose estimation of neural networks
%1. describe what researchers are already doing. Use the term above specifically (or explain synonyms)
%2. highlight the limitation / shortcoming
%3. foreshadowing: what would be the next step here (that will be taken in this contribution). or does your contribution build on it?

% zero-shot segmentation:
%UCN, UOIS, INSTR (stereo)
%currently known as segment anything
The development of deep neural networks for zero-shot object segmentation went through different stages. The problem in the context of robotic grasping was known as \textit{category-agnostic} and \textit{unseen object} instance segmentation. SD-Mask-RCNN~\cite{sdmaskrcnn}, UOIS~\cite{uois}, MSMFormer~\cite{msmformer}, and INSTR~\cite{instr} introduced different network architectures for RGB, depth, or RGB-D modalities.
Recently, the Segment Anything Model (SAM)~\cite{segment_anything} introduced the promptable segmentation task, where a prompt can be a set of points, a bounding box, or even text. With a prompt grid covering the input image, SAM can even be applied in a zero-shot "segment everything" mode, returning a full \mbox{(over-)}segmentation of the scene.
In our approach, we use SAM as the segmentation stage.
%Models got more generic with the release of Segment Anything Model~\cite{} for zero-shot object segmentation. Any of these models can be used for the first stage depending on the application requirement (RGB, Depth or RGB-D). We use SAM for the segmentation stage in our evaluation.

% figure* tends to shift to the next page, so bring it early
\begin{figure*}
    \vspace{0.3cm}
    \centering
    \includegraphics[width=0.99\textwidth, trim={100, 200, 100, 175}, clip]{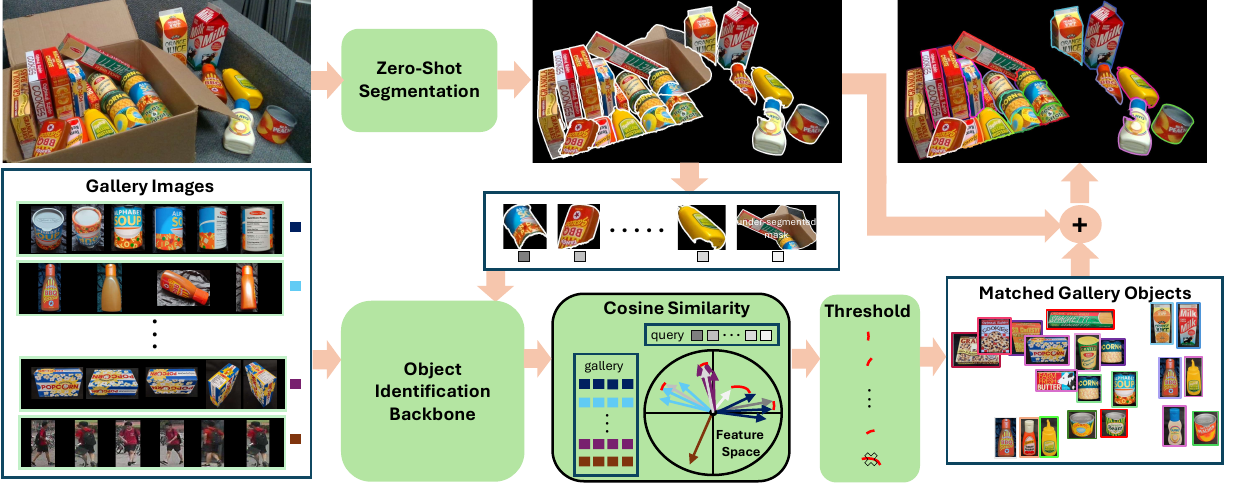}
    \caption{
    Full unseen object detector pipeline.
    A zero-shot segmentation method removes the background and produces object segments, which may be over- or undersegmented and can overlap.
    Features are extracted by the identification backbone on both gallery images and segments.
    After finding closest matches in feature space using cosine similarity, badly or un-matched items are rejected by a thresholding operation.
    Finally, the matches can be used to construct a full instance segmentation of the scene.}
    \label{fig:pipeline}
\end{figure*}

% object identification
Classification can be either object-specific of category-specific. Here we focus on the object identification/matching which is object-specific. The object identification problem falls under the image retrieval problem. This area was lacking datasets that are big enough to achieve practical performance until the release of the ARMBench dataset~\cite{armbench}.
The dataset contains multiple images both for query and for gallery (set of object's pre-captured images). The evaluation is carried out in two situations: First using a single image of the object in a bin before picking (pre-pick) and second using multiple (3) images after grasping the object in isolation (called post-pick). The post-pick includes the pre-pick image.
RoboLLM~\cite{robollm} achieved great accuracy on the ARMBench dataset. It uses a variant of the BEiT3~\cite{beit3} model followed a Multi-Layer Perceptron (MLP) for feature aggregation. A drawback of using MLPs is that the network can only use a fixed number of query-images. RoboLLM trained two different models to carry out the evaluation for pre-pick and post-pick situations.
In contrast, our approach is flexible regarding the backbone choice and allows any number of query or gallery images. It can also match multiple query objects at once.
%Our main concern is this paper is how to train any backbone model for object identification on the ArmBench dataset without having any fixed size layer and without a specific number of input images.

% few-shot learning
% fewsol
%few-shot learning. The goal of object identification is to match the query to the gallery objects, while the goal of few-shot learning is to classify the query objects on the category level. FewSol \cite{} introduced a dataset for few-shot learning to benchmark with different few-shot learning networks.

% template matching:
% difference between object identification and template matching
% template matching: does both segmentation and classification
% object identification does the image association only
% old methods searching all over the images - not flexible
% DTOID

Deep template matching is an analogous approach to 2D segmentation of unseen objects. HU et al. \cite{template_matching_Hu} and DTOID \cite{dtoid} introduced different approaches to detect and segment objects using only a few gallery images.
While architecturally pleasing, combining segmentation and identification in a single model is a hurdle that makes models harder to develop and, most importantly, harder to analyze shortcomings.
In contrast, depending on a separate zero-shot segmentation module allows us to leverage modern foundation models for this task and results in easy analysis of segmentation and classification performance.
%Unseen object instance segmentation depending on zero-shot segmentation and object identification is a more modern approach compared to deep template matching.

DoUnseen \cite{dounseen} and CNOS \cite{cnos} pipelines focus on the 2D segmentation of unseen objects. DoUnseen uses a variant of Mask R-CNN to extract object segments followed by a ViT model pre-trained on ImageNet for the object identification. CNOS uses SAM followed by DINOv2 model for the object identification. SAM-6D~\cite{sam6d} follows a similar scheme as CNOS for the 2D segmentation followed by a stage for 6D localization trained using the MegaPose~\cite{megapose} dataset. These three pipelines follow the scheme shown in Figure~\ref{fig:unseen_6d_pipelines} and can enhance their performance by replacing their object identification/matching models with our backbone as we surpass DINOv2 scores as shown in Section~\ref{sec:results}.

%The BOP challenge for 6D pose estimation in 2023 \cite{bop2022} introduced a new task for model-based 2D segmentation and 6D localization of unseen objects.
\section{Method}
\label{sec:method}

%Add sentences here: contextualize with regards to the new figure 2 (pipeline), stick to its terminology

In this section, we explain the centroid triplet loss and training details necessary for applying it efficiently on large-scale multi-query datasets such as ARMBench.
%We elaborate how we use the ARMBench dataset and what tricks we did to make the training feasible and faster.

\begin{figure}
    \centering
    \subfigure[Object X002W83UVZ]
    {
        \includegraphics[height=3cm]{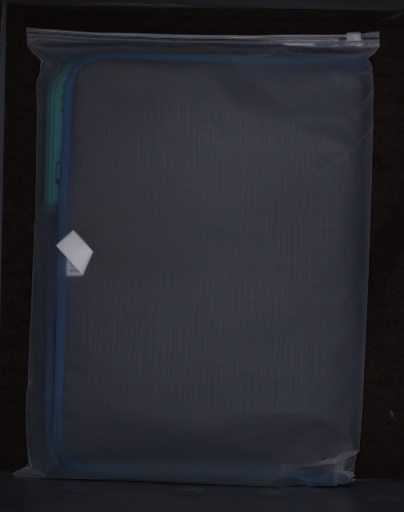}
        \includegraphics[height=3cm]{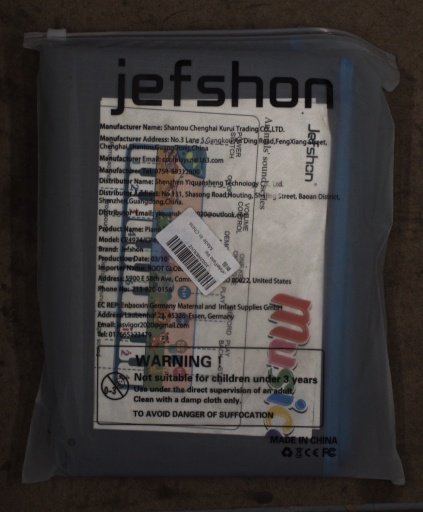}
        \includegraphics[height=3cm]{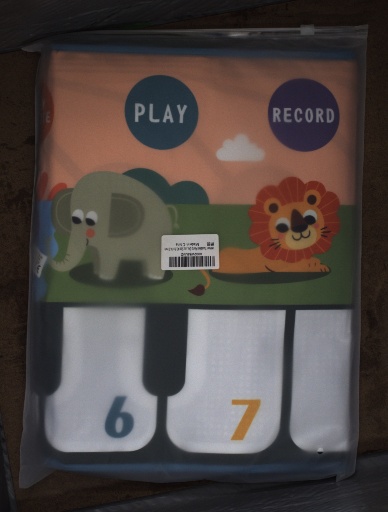}
    }
    \hspace{0.2cm}
    \subfigure[Object X0013DYNU7]
    {
        \includegraphics[height=3cm]{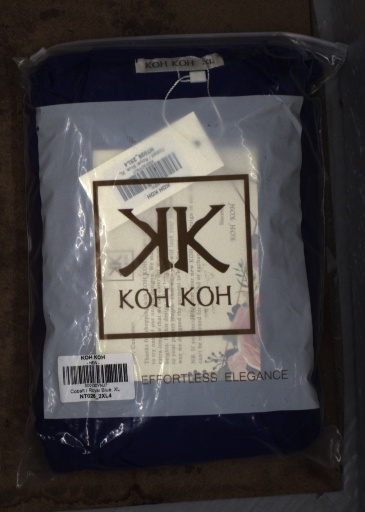}
        \includegraphics[height=3cm]{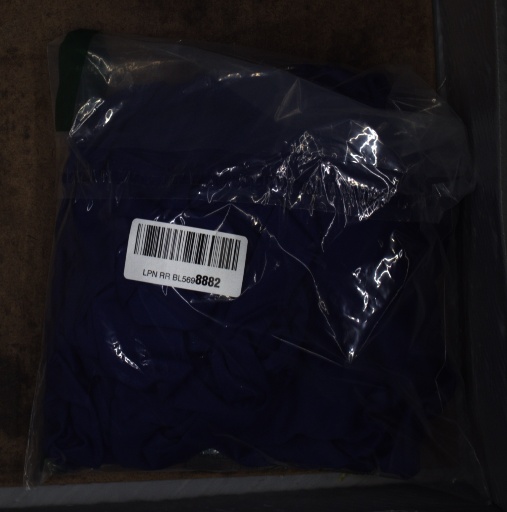}
        \includegraphics[height=3cm]{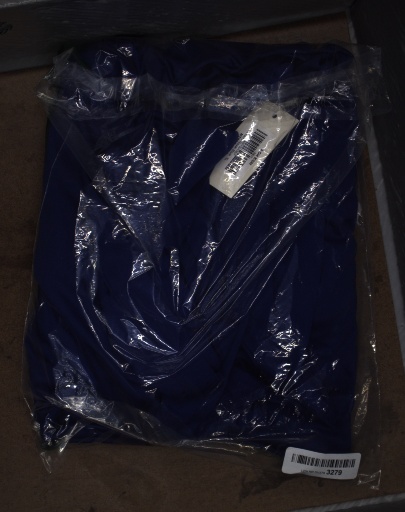}
    }
    \caption{
    Example objects from ARMBench that have gallery images differing largely in texture.
    Treating each image on its own as a possible match loses a valuable training signal---these belong together.
    Using CTL, we treat objects as a whole.
    %Different faces of gallery object might have different faces and texture from the back or the sides. If only one gallery image as the positive sample in the triplet loss, training data might be misleading. As this one gallery image might be of a different texture or color. Using CTL we solve this problem by using an object as a whole.
    }
    \label{fig:object_faces}
    \vspace{-0.5cm}
\end{figure}

\subsection{Centroid Triplet Loss for Object Identification}

%  Centroid Triplet Loss is the solution
How object images should be fed to a model for training is different to the related image retrieval and association tasks (e.g. person re-identification). In person re-identification datasets any of the gallery images could be used as a positive sample as humans tend to look similar from different perspectives.
But in case of objects for robotic grasping, different faces of an object can look very different from each other. As shown in Figure~\ref{fig:object_faces}, the front of an object can be of different color and texture from the back or the side.
Training on single views would discard the relationship to the other views, which is a valuable training signal.
%Using only a single image of one side of an object would feed a lot of misleading data to the model.
Therefore, an object should be fed to the model as a whole. A solution for this is to use the centroid triplet loss (CTL) as inspired by~\cite{unreasonable_ctl}. As shown in \cref{fig:ctl}, all images of an object (both query or gallery) are fed to a backbone and the resulting features are aggregated to their mean. The triplet loss is then calculated as follows:
\begin{equation}
\mathcal{L}_{\text{triplet}} = \max \left( \smash[b]{\underbrace{\lVert C_a - C_p \rVert_2}_{\textcolor{green!50!black}{\mathcal{L}_+}}} - \smash[b]{\underbrace{\lVert C_a - C_n \rVert_2}_{\textcolor{red!50!black}{\mathcal{L}_-}}} + \alpha, 0 \right),
\vphantom{\underbrace{\lVert C_a - C_p \rVert_2}_{\textcolor{green!50!black}{\mathcal{L}_+}}}
\end{equation}
where $C_a$, $C_p$ and $C_n$ are the centroids of the query object, the positive gallery object and the negative gallery object, respectively.
%In our case, the $\mathcal{l}^2$ norm is used for the distance.
Finally, $\alpha$ defines the margin of this loss function.

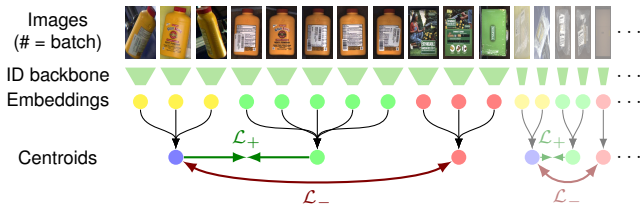
\begin{figure}
  \centering
  \newlength\twidth\setlength\twidth{0.4cm}
  \newlength\stwidth\setlength\stwidth{0.2cm}
  \newlength\theight\setlength\theight{0.7cm}
  \begin{tikzpicture}[
    font=\sffamily\scriptsize,
    second/.style={opacity=0.5}
  ]
      \matrix (images) [
        %label={below:Gallery \#1},
        inner sep=5pt,
        column sep=2pt,
        matrix of nodes,
        nodes={inner sep=0pt,anchor=center},
    ] {
        \includegraphics[width=\twidth,height=\theight,angle=180]{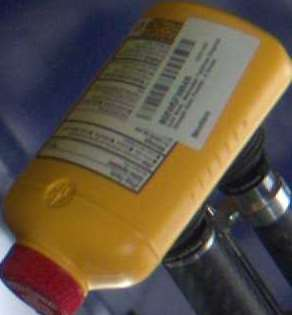} &
        \includegraphics[width=\theight,height=\twidth,angle=-90]{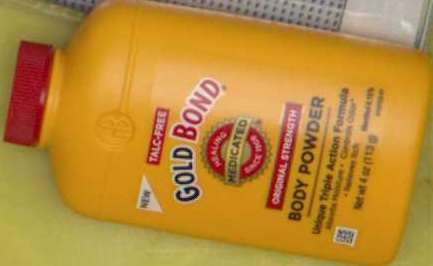} &
        \includegraphics[height=\theight,width=\twidth,angle=180]{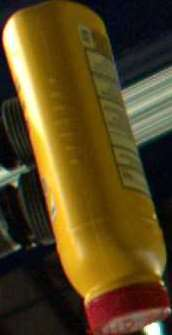} &
        \includegraphics[height=\theight,width=\twidth]{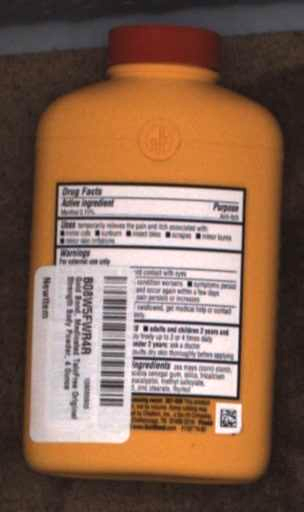} &
        \includegraphics[height=\theight,width=\twidth]{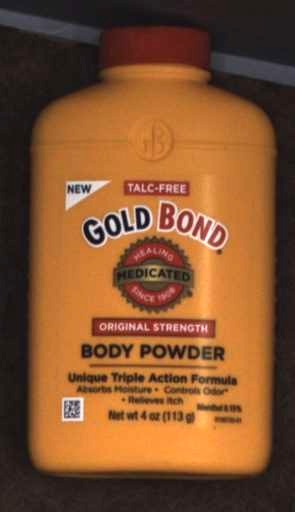} &
        \includegraphics[height=\theight,width=\twidth]{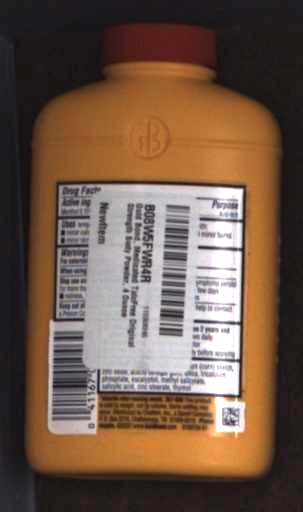} &
        \includegraphics[height=\theight,width=\twidth]{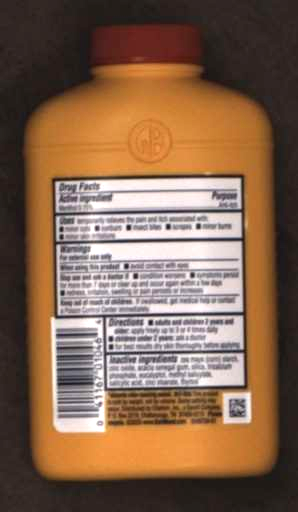} &
        \includegraphics[height=\theight,width=\twidth]{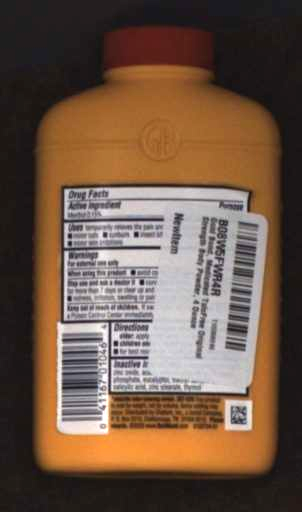} &
        \includegraphics[height=\theight,width=\twidth]{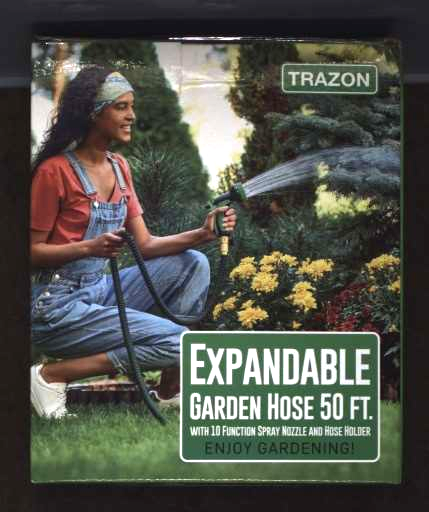} &
        \includegraphics[height=\theight,width=\twidth]{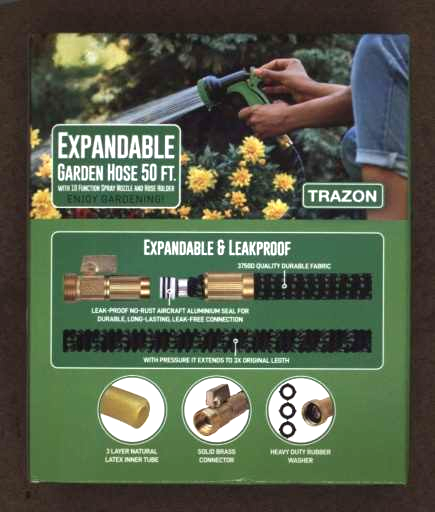} &
        \includegraphics[width=\theight,height=\twidth,angle=90]{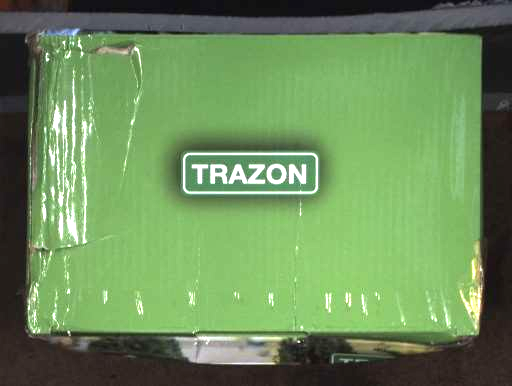} &
        |[second]| \includegraphics[width=\stwidth,height=\theight,angle=180]{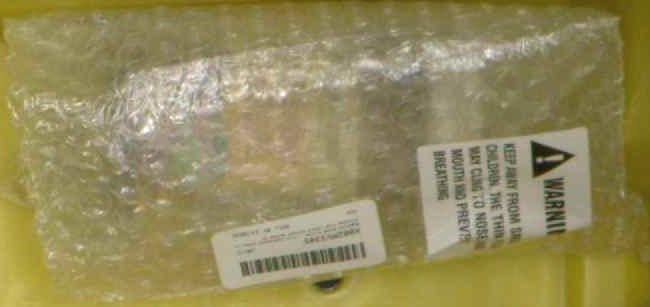} &
        |[second]| \includegraphics[height=\theight,width=\stwidth]{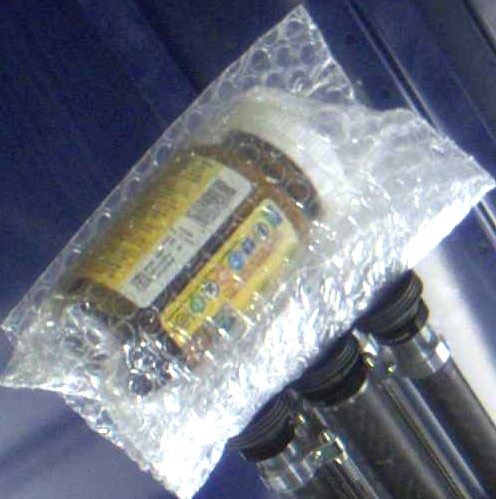} &
        |[second]| \includegraphics[height=\theight,width=\stwidth]{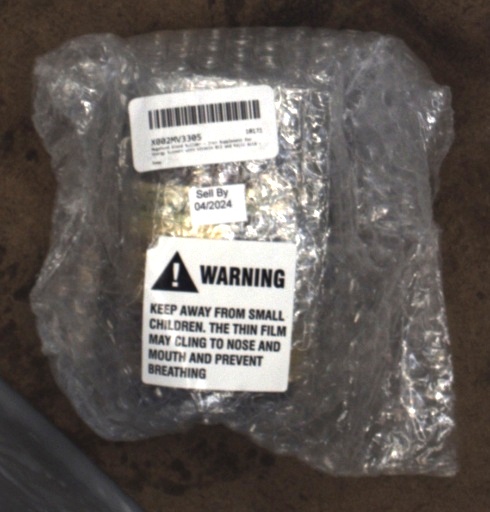} &
        |[second]| \includegraphics[height=\theight,width=\stwidth]{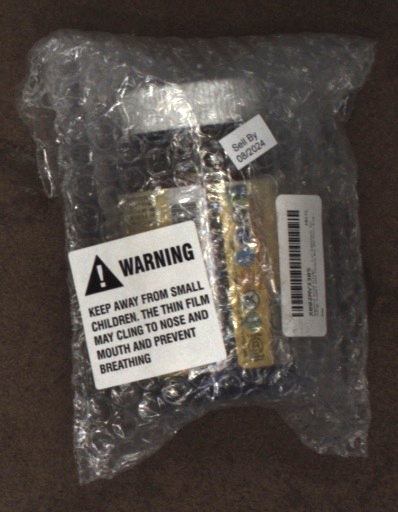} &
        |[second]| \includegraphics[height=\theight,width=\stwidth]{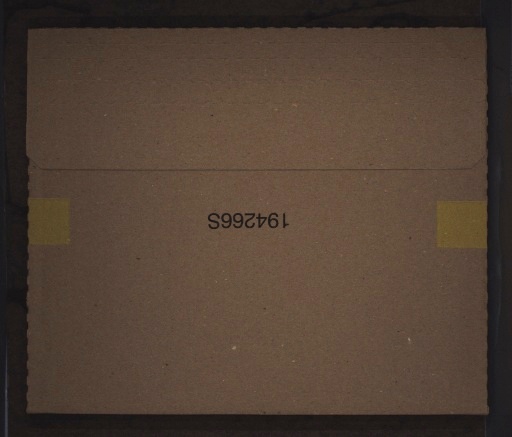} &
        $\dots$ \\
    };
    \foreach \i in {1,2,...,11} {
        \node[below=3pt, trapezium, trapezium angle=120, inner xsep=2pt,fill=anasgreen] (enc-\i) at (images-1-\i|-images-1-1.south){};
    }
    \foreach \i in {12,...,16} {
        \node[below=3pt, trapezium, trapezium angle=100, inner xsep=1pt,fill=anasgreen] (enc-\i) at (images-1-\i|-images-1-1.south){};
    }
    
    \foreach \i in {1,2,3} {
        \node[below=3pt of enc-\i,circle,fill=yellow!80,inner sep=2pt] (feat-\i) {};
    }
    \foreach \i in {4,...,8} {
        \node[below=3pt of enc-\i,circle,fill=green!50,inner sep=2pt] (feat-\i) {};
    }
    \foreach \i in {9,...,11} {
        \node[below=3pt of enc-\i,circle,fill=red!50,inner sep=2pt] (feat-\i) {};
    }

    \begin{scope}[second]
    \foreach \i in {12,13} {
        \node[below=3pt of enc-\i,circle,fill=yellow!80,inner sep=2pt] (feat-\i) {};
    }
    \foreach \i in {14,15} {
        \node[below=3pt of enc-\i,circle,fill=green!50,inner sep=2pt] (feat-\i) {};
    }
    \foreach \i in {16} {
        \node[below=3pt of enc-\i,circle,fill=red!50,inner sep=2pt] (feat-\i) {};
    }
    \end{scope}

    \node[below=15pt of feat-2,circle,fill=blue!50,inner sep=2pt] (q1) {};
    \draw[latex-,out=90,in=-90] (q1) edge (feat-1) edge (feat-2) edge (feat-3);

    \node[below=15pt of feat-6,circle,fill=green!50,inner sep=2pt] (p1) {};
    \foreach \i in {4,...,8} {
        \draw[latex-,out=90,in=-90] (p1) to (feat-\i);
    }

    \node[below=15pt of feat-10,circle,fill=red!50,inner sep=2pt] (n1) {};
    \foreach \i in {9,...,11} {
        \draw[latex-,out=90,in=-90] (n1) to (feat-\i);
    }

    \coordinate (midpos) at ($(q1)!0.5!(p1)$);
    \draw[-latex,thick,green!50!black] (q1) -- (midpos);
    \draw[-latex,thick,green!50!black] (p1) -- (midpos);
    \node[green!50!black,above=0cm of midpos] {$\mathcal{L}_+$};

    \draw[bend right,latex-latex,thick,red!50!black] (q1) to[out=-30,in=-150,looseness=0.5,edge label'={$\mathcal{L}_-$}] (n1);

    \begin{scope}[second,transparency group]
        \node[below=15pt,circle,fill=blue!50,inner sep=2pt] (q2) at ($(feat-12.south)!0.5!(feat-13.south)$) {};
        \draw[latex-,out=90,in=-90] (q2) edge (feat-12) edge (feat-13);

        \node[below=15pt,circle,fill=green!50,inner sep=2pt] (p2) at ($(feat-14.south)!0.5!(feat-15.south)$) {};
        \foreach \i in {14,15} {
            \draw[latex-,out=90,in=-90] (p2) to (feat-\i);
        }

        \node[below=15pt of feat-16,circle,fill=red!50,inner sep=2pt] (n2) {};
        \foreach \i in {16} {
            \draw[latex-,out=90,in=-90] (n2) to (feat-\i);
        }

        \coordinate (midpos) at ($(q2)!0.5!(p2)$);
        \draw[-latex,green!50!black] (q2) -- (midpos);
        \draw[-latex,green!50!black] (p2) -- (midpos);
        \node[green!50!black,above=0cm of midpos] {$\mathcal{L}_+$};

        \draw[bend right,latex-latex,thick,red!50!black] (q2) to[out=-50,in=-130,looseness=1.2,edge label'={$\mathcal{L}_-$}] (n2);
    \end{scope}

    \node at ({images-1-17}|-{enc-1}) {$\dots$};
    \node at ({images-1-17}|-{feat-1}) {$\dots$};
    \node at ({images-1-17}|-{q1}) {$\dots$};

    \node[left=0pt of images,align=center] (imglabel) {Images\\(\# = batch)};
    \node[anchor=north,inner ysep=0pt] at (imglabel|-enc-1.north) {ID backbone};
    \node at (imglabel|-feat-1) {Embeddings};
    \node at (imglabel|-q1) {Centroids};
  \end{tikzpicture}
  \vspace{-4ex}
  \caption{Batch computations during training. Each batch is filled
  with triplets until the batch size is reached.
  Backbone feature vectors are extracted in batched fashion.
  The resulting features are then aggregated to their corresponding centroid using an \textit{index-add} operation. Finally, the losses are applied.
  }
  \label{fig:batching}
\end{figure}

Since the number of images per object is not constant for both query and galleries, efficient training requires careful batch management.
To keep an optimal batch size $B$, we greedily keep adding triplets of the query as well as positive and negative galleries with their corresponding images to the batch until $B$ is reached (see \cref{fig:batching}).
The embeddings are then extracted in typical batched fashion over all images.
Crucially, recording a corresponding centroid index $i_c$ for each image
allows efficient batched summation and division, so that each object's centroid can be computed directly.
Finally, the centroid triplet loss can be applied for each triplet, also in batched fashion.
This method of batching and aggregation ensures optimal GPU utilization and scales to multi-GPU training by centrally pre-computing the batch splits for each epoch and then dividing the total number of batches across GPUs.

% object batches
%The number of images per object in the ARMBench dataset is not fixed. Each object's gallery contains between 1 and 6 images. This irregularity results in uneven batch sizes, which complicates feature aggregation.
%%To be able to calculate the centroids (the mean) directly on the GPU, the number of query or gallery images per object in a training batch must be fixed. 
%To solve this issue we use n-folds/repetition of the images to force all objects to have exactly 6 images as shown in \cref{tab:fold_repeat}.

%\begin{table}
%    \caption{Repetition modes for gallery images}
%    %\caption{To make the number of gallery images fixed during training we use folding/repetition.}
%    \label{tab:fold_repeat}
%    \centering
%    \begin{tabular}{cc}
%    \toprule
%    \# gallery images & Repetition \\
%    \midrule
%    1 &  6-fold\\
%    2 &  3-fold\\
%    3 &  2-fold\\
%    4 &  randomly repeat 2 images\\
%    5 &  randomly repeat 1 images\\
%    6 &  identity\\
%    \bottomrule
%    \end{tabular}
%\end{table}

%The folding will not change the centroids of the objects but repeating will. The repeating will cause a shift in the centroid towards the randomly chosen images. However, this shift should not be significant based on the fact that these images are of the same object and the repeated element is randomly chosen. Also in comparison to the effect this trick has on efficiency, this should still be acceptable. This method shortened the training time by 1/5 of the time required without it, which could not have been reached without CTL.

% threshold
During inference, we need a method to match the query centroid $q$ and the gallery images. For this, we use the cosine similarity score
\begin{equation}
    s(x,y) = 1 - \frac{x \cdot y}{\lVert x \rVert_2 \lVert y \rVert_2}  \label{eq:cosine}
\end{equation}
and select the most similar gallery entry $g$ with maximum $s(q,g)$.
We discard matches with $s < \theta$, where $\theta$ is a hyperparameter.
This will not only omit bad association, but will particularly stop matching non-existing objects in the query or the scene.
For evaluation on the ARMBench object test set this is not necessary, but the thresholding plays an important role when evaluating the whole unseen object instance segmentation pipeline.

\subsection{2D Segmentation of Unseen Objects}

\Cref{fig:pipeline} shows our full pipeline for the object detection.

For the zero-shot segmentation we use a combination of Mask R-CNN from \cite{maskrcnn} and SAM \cite{segment_anything}. The Mask R-CNN is only responsible for background extraction. The Mask R-CNN is trained on NVIDIA Falling Things dataset~\cite{falling_things_dataset} with only one class representing any warehouse object. The image is then segmented by SAM and any segmentation masks belonging to the background (according to Mask R-CNN) are discarded. We note this method as (Mask R-CNN+SAM) in our upcoming evaluation. As SAM is prompt-based, SAM is prompted with a grid of points to segment all object in the image. This can lead to many over- and under-segmented masks that even overlap. We depend on the thresholding operation defined above to filter the over- and under-segmented masks as shown in \cref{sec:eval}.

\section{Experiments}
\label{sec:results}\label{sec:eval}

In this section, we carry out two evaluations: first on the test set of the ARMBench object identification dataset~\cite{armbench} and, second, on the HOPE dataset~\cite{hope_dataset}.

\begin{table*}[h]

\vspace{0.3cm}
\centering
\begin{threeparttable}
\caption{
Evaluation on ArmBench object identification test set
}
\label{tab:armbech_eval}
\setlength{\tabcolsep}{3pt}
\begin{tabular}{l rc@{\hspace{2em}} cc@{\hspace{2em}} cc@{\hspace{2em}} cc}
\toprule
 & & & \multicolumn{2}{@{}c@{\hspace{2em}}}{Recall@1} & \multicolumn{2}{@{}c@{\hspace{2em}}}{Recall@2} & \multicolumn{2}{@{}c}{Recall@3} \\
 \cmidrule (r{2em}) {4-5}
\cmidrule (r{2em}) {6-7}
\cmidrule (r{2pt}) {8-9}
 Method & \# query images & Trained on & pre & post & pre & post & pre & post \\
\midrule
ResNet50-RMAC \cite{resnet_rmac}  & any & ImageNet  & 71.7 & 72.2 & 81.9 & 82.9 & 87.2 & 88.2 \\
DINO-ViTS \cite{DINO_ViTS}        & any & ImageNet & 77.2 & 79.5 & 87.3 & 89.4 & 91.6 & 93.5 \\
%dinov2_vitb14_reg
DINO-V2 \cite{dinov2_registers}   & any & ImageNet & 72.3 & 75.1 & 84.2 & 87.5 & 89.7 & 92.6 \\
\midrule
ViT-b-16-CTL-instance (ours)      & any & ArmBench & 97.2 & \textbf{99.3} & 97.8 & 99.4 & 98.3 & 99.6 \\
ViT-b-16-CTL-centroid (ours)      & any & ArmBench & 97.2 & 98.6 & \textbf{99.0} & \textbf{99.5} & \textbf{99.4} & \textbf{99.7} \\

Resnet-50-CTL-instance (ours)     & any & ArmBench &  88.4 & 97.0 & 90.8 & 98.0 & 92.6 & 98.5 \\
Resnet-50-CTL-centroid (ours)     & any & ArmBench & 86.9 & 94.3 & 94.5 & 98.1 & 96.9 & 99.0 \\

\midrule
RoboLLM \cite{robollm}            & 1 or 3 & ArmBench & \textbf{97.8} & 98.0 & 97.9 & 98.1 & 98.0 & 98.2 \\
\bottomrule

\hline
\end{tabular}
We report the Recall@k metric for $k \in {1,2,3}$ for the pre-pick and post-pick situations.
Our ViT model scores the highest accuracy on the test set for the multi-query (post-pick) situation.
Using the closest gallery object instance (shown as "-instance") gives higher accuracy than searching for the closest object centroid ("-centroid") and
is thus recommended for inference.
%Using our model instead of a model pretrained on ImageNet (as is commonly done) can increase performance of
%any unseen object detection pipeline.
%Our model scores higher than DINOv2 which shows that our model is able to enhance the pre-mentioned pipelines for unseen object as they depend on DINOv2 weights.
\end{threeparttable}
\end{table*}

\begin{table*}
    \setlength{\tabcolsep}{3pt}
    \centering
    \begin{threeparttable}
    \caption{
    Evaluation on the HOPE validation set (bounding boxes)
    }
    \label{tab:eval_hope_bbox}
    \begin{tabular}{l@{\hspace{2em}}c@{\hspace{2em}} *{10}{c}}
        \toprule
        Method & Training on HOPE & AP & AP50 & AP75 & AP$_M$ & AP$_L$ & AR$_1$ & AR$_{10}$ & AR$_{100}$ & AR$_M$ & AR$_L$\\
        \midrule
        Mask R-CNN & HOPE-Video
        & 0.196 & 0.377 & 0.206 & 0.090 & 0.200 & 0.231 & 0.294 & 0.294 & 0.087 & 0.298 \\
        Bonani et al. \cite{learn_from_sam} & Meshes + unlabeled HOPE-Video
        & 0.338 & \textbf{0.552} & 0.364 & 0.172 & \textbf{0.380} & \textbf{0.387} & \textbf{0.452} & \textbf{0.452} & \textbf{0.220} & \textbf{0.457} \\
        \midrule
        SAM + DINOv2  & none
        & 0.316 & 0.431 & 0.346 & \textbf{0.180} & 0.317 & 0.339 & 0.383 & 0.383 & 0.217 & 0.386 \\
        SAM + ViT-CTL (Ours)  & none
        & \textbf{0.349} & 0.494 & \textbf{0.377} & 0.105 & 0.367 & 0.384 & 0.438 & 0.438 & 0.148 & 0.447 \\
        \midrule
        DINOv2 (GT masks)  & none
        & 0.581 & 0.581 & 0.581 & 0.277 & 0.587 & 0.582 & 0.671 & 0.671 & 0.275 & 0.679 \\
        Ours (GT masks) & none
        & 0.740 & 0.740 & 0.740 & 0.277 & 0.755 & 0.680 & 0.776 & 0.776 & 0.275 & 0.791 \\
        \bottomrule
    \end{tabular}
    We report the standard COCO metrics. Note: AP$_S$ and AR$_S$ are not applicable, since the dataset does not contain "small" segments.
    Highest numbers (except for ground truth baselines) are highlighted in bold. The segmentation method (Mask R-CNN+SAM) is simply denoted (SAM) in this table.
    %Evaluation of our 2D segmentation of unseen objects pipeline against trained method on the HOPE datasets using COCO metrics.
    \end{threeparttable}
    %Note the low detection of hope is close to the non multi view as their original paper with DOPE around 0.15
\end{table*}

\begin{table*}
    \setlength{\tabcolsep}{3pt}
    \centering
    \begin{threeparttable}
    \caption{
    Evaluation on the HOPE validation set (segmentation)
    }
    \label{tab:eval_hope_segm}
    \begin{tabular}{l@{\hspace{2em}}c@{\hspace{2em}} *{12}{c}}
        \toprule
        Method & Training on HOPE & AP & AP50 & AP75 & AP$_M$ & AP$_L$ & AR$_1$ & AR$_{10}$ & AR$_{100}$ & AR$_M$ & AR$_L$\\
        \midrule
        Mask R-CNN & HOPE-Video
        & 0.182 & 0.354 & 0.188 & 0.031 & 0.186 & 0.212 & 0.272 & 0.272 & 0.030 & 0.279 \\
        Bonani et al. \cite{learn_from_sam} & Meshes + unlabeled HOPE-Video
        & 0.333 & \textbf{0.564} & 0.378 & \textbf{0.202} & 0.373 & 0.371 & 0.434 & 0.434 & \textbf{0.240} & 0.441 \\
        \midrule
        SAM + DINOv2  & none
        & 0.337 & 0.436 & 0.371 & 0.171 & 0.340 & 0.359 & 0.405 & 0.405 & 0.217 & 0.409 \\
        SAM + ViT-CTL (Ours)     & none
        & \textbf{0.374} & 0.520 & \textbf{0.403} & 0.099 & \textbf{0.395} & \textbf{0.405} & \textbf{0.462} & \textbf{0.462} & 0.135 & \textbf{0.472} \\
        \midrule
        DINOv2 (GT masks)  & none
        & 0.581 & 0.581 & 0.581 & 0.277 & 0.587 & 0.582 & 0.671 & 0.671 & 0.275 & 0.679 \\
        ViT-CTL (GT masks) & none
        & 0.740 & 0.740 & 0.740 & 0.277 & 0.755 & 0.680 & 0.776 & 0.776 & 0.275 & 0.791 \\
        \bottomrule
    \end{tabular}
    We report the standard COCO metrics. Note: AP$_S$ and AR$_S$ are not applicable, since the dataset does not contain "small" segments.
    Highest numbers (except for ground truth baselines) are highlighted in bold.
    The segmentation method (Mask R-CNN+SAM) is simply denoted (SAM) in this table.
    %Evaluation of our 2D segmentation of unseen objects pipeline against trained method on the HOPE datasets using COCO metrics.
    \end{threeparttable}
    %Note the low detection of hope is close to the non multi view as their original paper with DOPE around 0.15
\end{table*}

\begin{figure}
  \centering
  \subfigure[segmentation output]{\includegraphics[scale=0.49]{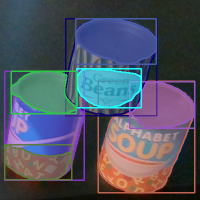}}\quad
  \subfigure[after classification]{\includegraphics[scale=0.49]{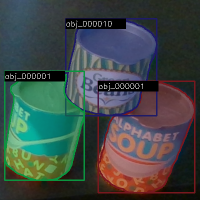}}
  \caption{Filtration of over/under-segmented masks by our object identification model. The filtration also handles removing of background segments and objects in gallery that are not present in the image.}
  \label{fig:filter}
\end{figure}

% \begin{figure*}
%     \vspace{0.2cm}
%     \centering
%     \includegraphics[width=0.45\textwidth]{figures/hope_eval_outputs/8.png}
%     \includegraphics[width=0.45\textwidth]{figures/hope_eval_outputs/2.png}
%     \\
%     \vspace{0.1cm}
%     \includegraphics[width=0.45\textwidth]{figures/hope_eval_outputs/3.png}
%     \includegraphics[width=0.45\textwidth]{figures/hope_eval_outputs/9.png}
%     \caption{Qualitative examples of segmentation and identification on the HOPE dataset.}
%     \label{fig:hope_eval_outputs}
% \end{figure*}

\begin{figure*}
    \vspace{0.2cm}
    \centering
    \subfigure[using GT masks]
    {
        \includegraphics[width=0.24\textwidth]{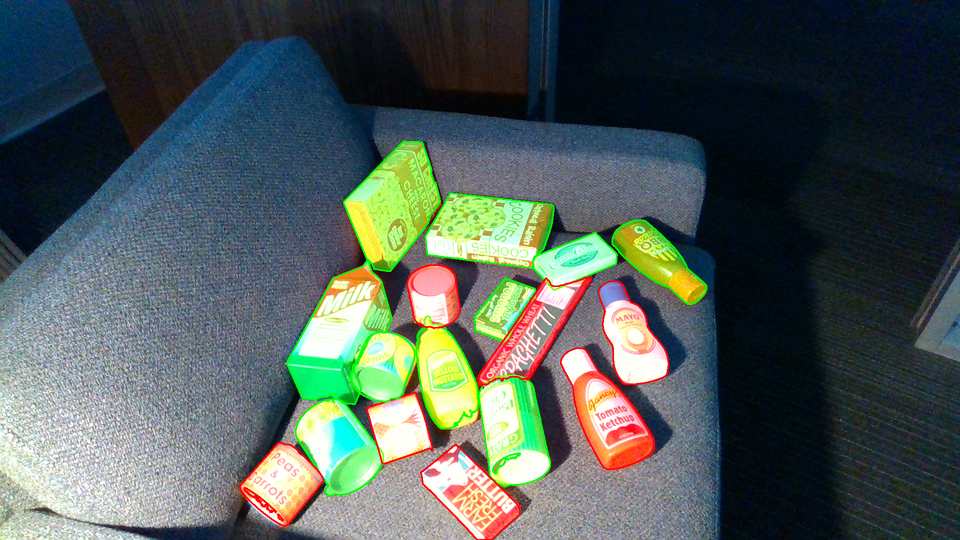}
        \includegraphics[width=0.24\textwidth]{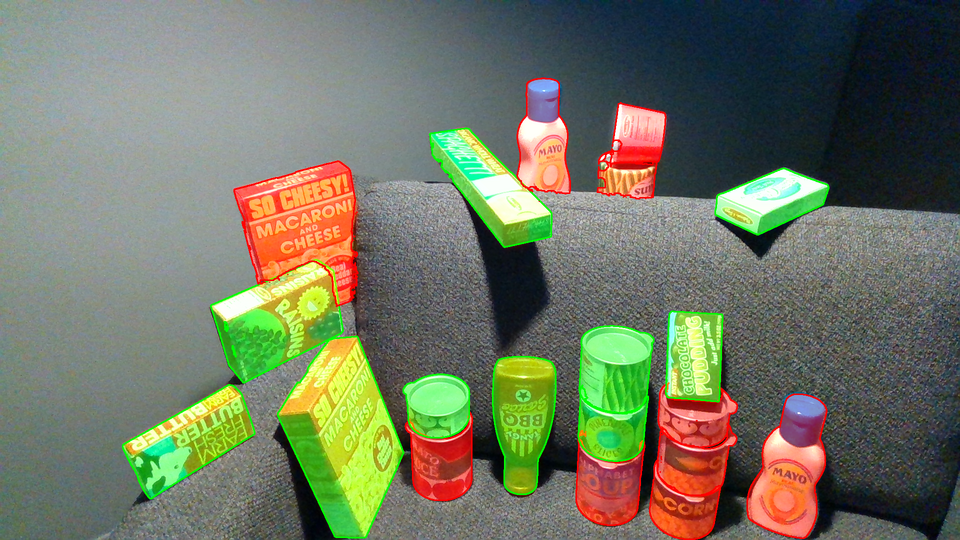}
        \includegraphics[width=0.24\textwidth]{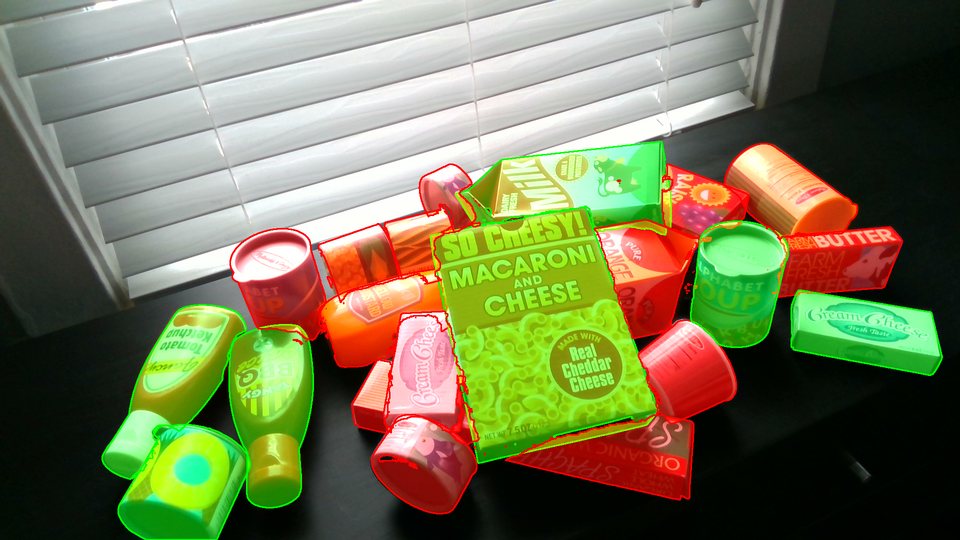}
        \includegraphics[width=0.24\textwidth]{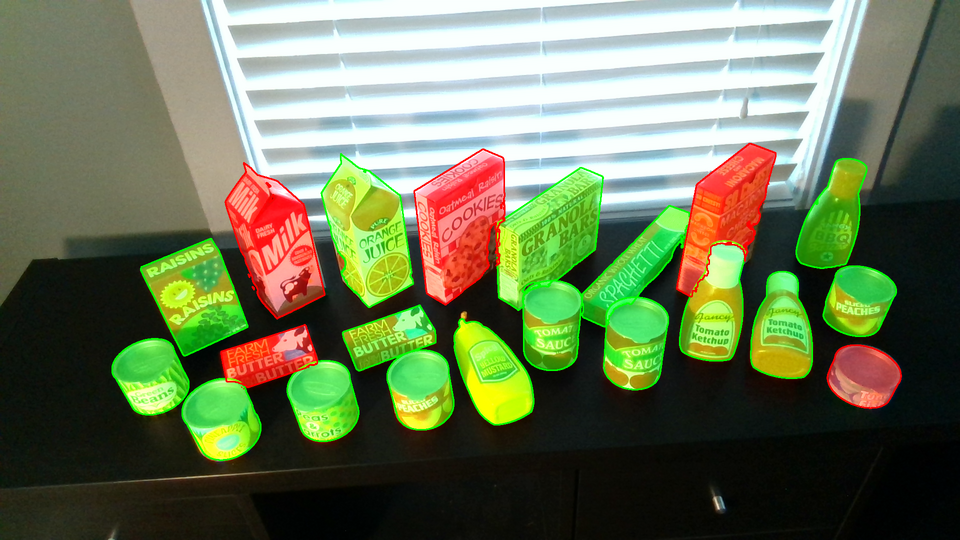}
    }\\
    \subfigure[using Mask R-CNN+SAM]
    {
        \includegraphics[width=0.24\textwidth]{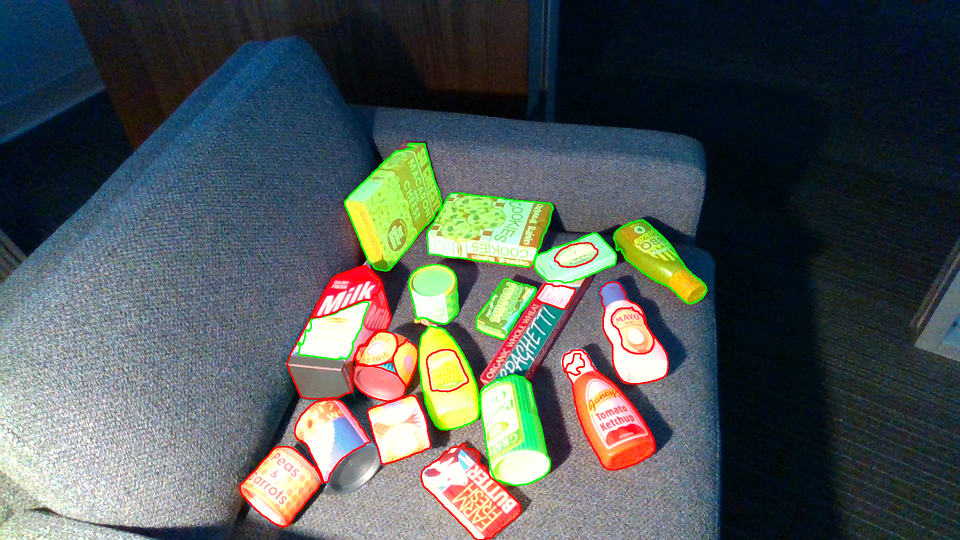}
        \includegraphics[width=0.24\textwidth]{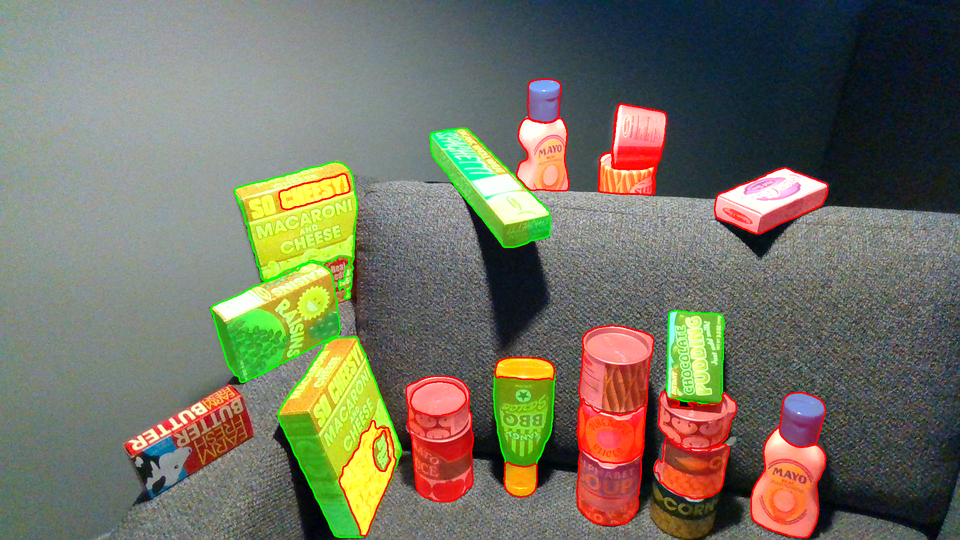}
        \includegraphics[width=0.24\textwidth]{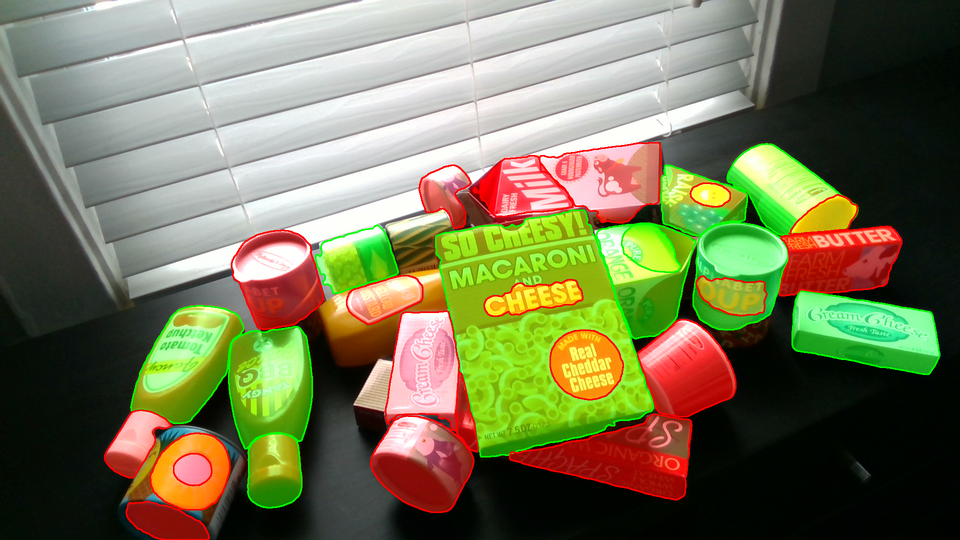}
        \includegraphics[width=0.24\textwidth]{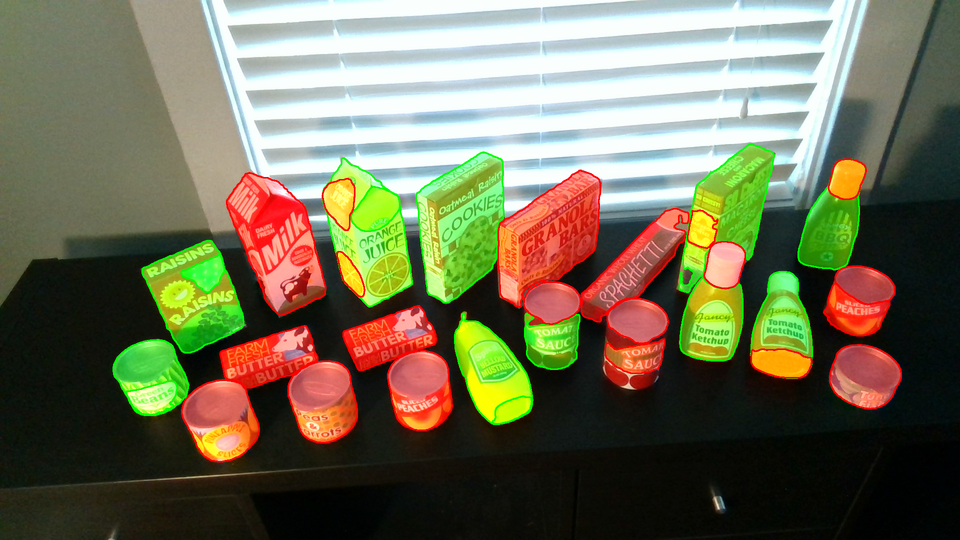}
    }
    \caption{Qualitative examples of segmentation and identification on the HOPE dataset. Green color represents objects that have been segmented and identified correctly. Red color represents segments that are identified as a wrong class.}
    \label{fig:hope_eval_outputs}
\end{figure*}

\subsection{Object Identification on ARMBench} %Evaluation of the object identification model using the ARMBench dataset

% Which dataset
To train a model for object identification a dataset of thousands of objects is required. 
This is what the object identification section of the ARMBench dataset \cite{armbench} provides. It contains 190K gallery objects (Reference images) with multiple images for each gallery object and contains 235K query scene (Picks) also with multiple images for each query object. This large number of objects provides enough data to train a model that is able to generalize to new objects at inference. From the 235K query scenes, 50k are used for test.
%This large test set would assure the real world capabilities of the tested methods.

We follow the evaluation protocol in \cite{robollm} and report the Recall@k metric for $k \in {1,2,3}$.
We also differentiate between the pre-pick and post-pick situations, where pre-pick is the captured image of the object inside the bin before the robot arm grasps it. The post-pick situation includes the pre-pick image and other images of the object while it is grasped by the robot.

% model
Our approach is generic to the actual backbone architecture. In our evaluation, we focus on ResNet~\cite{resnet} and ViT~\cite{vit}.
Both models were pretrained on ImageNet. For ResNet we select ResNet-50 and train for 100 epochs (1 week) on one A100 GPU with a learning rate of 1e-3, 1e-4 and 1-e5 for 40, 30 and 30 epochs, respectively.
% all query images: 1)40X0.001 2)10*0.0001 3)10*0.003
% one query images: 4)10*0.007 5)10*0.014 6)6*0.014 7) 20*0.1+10*0.05
For ViT, we use a ViT-b-16 instance and train for 100 epochs on eight A100 GPUs (approximately three days) using SGD with a learning rate of 0.05. Following Kumar et al.~\cite{finetuning_vit}, we freeze the first layer of the ViT model. To improve robustness, we train an additional 100 epochs with data augmentation (TrivialAugment~\cite{trivial_aug}).

To prepare the model for single query images, we select the first query image of ARMBench instead of the query centroid during training with a random chance of 50\%.

Table \ref{tab:armbech_eval} shows the result of our trained ViT and ResNet. We compare our model to ResNet-50-RMAC \cite{resnet_rmac} and DINO-ViTS \cite{DINO_ViTS} which are trained on ImageNet as evaluated in the ARMBench dataset paper \cite{armbench}. Our variant with ViT scores the highest post-pick results among all methods. It is also worth noting that the accuracy of the ViT increases from 97.2\,\% to 99.3\,\% when using multiple query images (post-pick) instead of a single image (pre-pick). This shows the accuracy can increase with the addition of more query images, which is useful in applications where more query images are collected automatically.

\subsection{2D Segmentation of Unseen Objects}

% Evaluation dataset
%HOPE is easy for segmentation, challenging for classification
In this section we evaluate the whole pipeline on the HOPE\footnotemark ~dataset (validation split). The reason we choose the HOPE dataset is that its objects look very similar, requiring fine-grained classification, which is challenging for an object identification model and tests the hard upper limits of the model. The evaluation is done using the COCO metrics for bounding box detection and segmentation. Similar to other datasets in the BOP format, the HOPE dataset offers modal (visible) and amodal (full) masks of the objects in 2D. Here we evaluate with modal masks. The gallery images are taken manually for each object covering all the unique faces of each objects. The gallery images are then augmented by rotating each image multiples of 45 degree.

\footnotetext{Evaluation was conducted using HOPEv1~\cite{hope_dataset}. A more recent version, HOPEv2, was subsequently released in May 2024 for the BOP challenge 2024.}

We offer two baselines: First, we train a Mask R-CNN model in supervised fashion for instance segmentation using the HOPE-video dataset \cite{hope_video}.
We note that this baseline requires an annotated training set.
The second baseline considers the case where object models and unlabeled real data is available. For this unlabeled data the HOPE-Video dataset is used but the labels are disregarded and not used. For this second baseline, we apply the method of Bonani et al.~\cite{learn_from_sam}, which trains a semantic segmentation network in supervised fashion on synthetic data, and uses SAM to regularize the network's output on the unlabeled real data.
We note that this method produces a semantic segmentation, which is then converted to instance segmentation by finding connected components. This adaptation is obviously sub-optimal when objects of the same class overlap.

We evaluate our pipeline for 2D segmentation of unseen objects once with a pre-trained DINOv2 backbone and once with our object identification model.
Each backbone is again evaluated twice: Once with the full pipeline, and once with ground truth segment masks. The evaluation with the ground truth masks helps estimate the contribution of the zero-shot segmentation and our object identification model to performance. For these evaluations we use a cosine similarity score threshold of $\theta=0.6$. As described in \cref{sec:method} (\cref{eq:cosine}), if score of the best match is lower than $\theta$ we discard this query segment as it might be an object that is not present in the gallery or an over-/under-segmentation.

\cref{tab:eval_hope_bbox,tab:eval_hope_segm} show quantitative results with metrics calculated on bounding boxes and segmentation masks, respectively.
Interestingly, our method clearly outperforms Mask R-CNN, which was trained in supervised fashion. This may indicate that the (comparably small) domain shift between HOPE and HOPE-Video is difficult for this purely supervised method to overcome.
Furthermore, the method of Bonani et al.~\cite{learn_from_sam} also beats Mask R-CNN, showing the usefulness of synthetic data generated from the object models, even in the absence of annotations on real training data.
Our full pipeline matches and even surpasses this performance in most metrics, which is highly interesting as this baseline has access to object models and unlabeled real data.
Finally, we can see that our object identification model surpasses DINOv2 across all COCO metrics.
When removing the effects of segmentation \& segment filtering by using ground truth masks, our model scores an AP of 0.740 against DINOv2 with an AP of 0.581 making our model score 27.4\% higher than DINOv2.
This supports our claim that our model can enhance pipelines for unseen object detection that are using DINOv2 such as CNOS~\cite{cnos} or SAM-6D~\cite{sam6d}.

Another interesting result is the filtration our object identification model can contribute. Figure \ref{fig:filter} (a) shows the output of the zero-shot segmentation with SAM. Figure \ref{fig:filter} (b) shows how our model filters the over-segmented masks as they score below the threshold.

Finally, we show exemplary qualitative results in \cref{fig:hope_eval_outputs}.

%The low score of HOPE is close to the score in their original paper with DOPE .....
\section{Conclusion}

% summarize:
% introduced CTL
% how to use CTL with ARMBench
% we achieved the highest score on ARMBench
% we surpass DINOv2
% multi query multi gallery at once
% Our model can take any number of query images
% the model has practical performance can be used in application

% limiations:
% objects that are similar - like hope
% a wide area for improvements

% future:
% new architecture - deeper multi-channel feature vector
% integrate and test model with 6D Pose pipelines

In this work we introduced how to use the centroid triplet loss for training models for object identification. We successfully showed that CTL scores the highest accuracy on the ARMBench test set. In particular, the trained model is able to process any number of query or gallery images. It is capable of matching multiple query object to multiple gallery object at once, shortening inference time.
The model performance surpasses DINOv2 on ARMBench and can serve as an improved backbone for 2D segmentation and 6D localization of unseen objects. The near-perfect score on ARMBench indicates that it can be used in real-world applications.

When combined with a generic zero-shot segmentation method such as SAM, the result is a complete segmentation pipeline. While performance on the challenging HOPE dataset is still limited in absolute terms, our pipeline beats several related methods that can access object information during training.
%The model is still limited in performance on the hard case of the HOPE dataset. This leaves a wide area for improvements.
%This leads us to think that a model with a deeper hierarchical representation might be a booster on the performance of object like the HOPE dataset.
Tighter integration with the segmentation module and intelligent proposal filtering might improve results further.

%\section*{APPENDIX} %In case I made an appendix

\section*{ACKNOWLEDGMENT}

This work is funded by the German Federal Ministry of Education and Research (BMBF) in the course of the Lamarr Institute for Machine Learning and Artificial Intelligence.

\bibliographystyle{IEEEtran}
\bibliography{IEEEabrv,./references.bib}

\begin{thebibliography}{10}
\providecommand{\url}[1]{#1}
\csname url@samestyle\endcsname
\providecommand{\newblock}{\relax}
\providecommand{\bibinfo}[2]{#2}
\providecommand{\BIBentrySTDinterwordspacing}{\spaceskip=0pt\relax}
\providecommand{\BIBentryALTinterwordstretchfactor}{4}
\providecommand{\BIBentryALTinterwordspacing}{\spaceskip=\fontdimen2\font plus
\BIBentryALTinterwordstretchfactor\fontdimen3\font minus \fontdimen4\font\relax}
\providecommand{\BIBforeignlanguage}[2]{{%
\expandafter\ifx\csname l@#1\endcsname\relax
\typeout{** WARNING: IEEEtran.bst: No hyphenation pattern has been}%
\typeout{** loaded for the language `#1'. Using the pattern for}%
\typeout{** the default language instead.}%
\else
\language=\csname l@#1\endcsname
\fi
#2}}
\providecommand{\BIBdecl}{\relax}
\BIBdecl

\bibitem{unreasonable_ctl}
M.~Wieczorek, B.~Rychalska, and J.~Dabrowski, ``On the unreasonable effectiveness of centroids in image retrieval,'' in \emph{28th International Conference on Neural Information Processing (ICONIP)}, ser. Lecture Notes in Computer Science, vol. 13111.\hskip 1em plus 0.5em minus 0.4em\relax Springer, 2021, pp. 212--223.

\bibitem{morrison2018cartman}
D.~Morrison, A.~W. Tow, M.~Mctaggart, R.~Smith, N.~Kelly-Boxall, S.~Wade-Mccue, J.~Erskine, R.~Grinover, A.~Gurman, T.~Hunn \emph{et~al.}, ``Cartman: The low-cost {Cartesian} manipulator that won the {Amazon Robotics Challenge},'' in \emph{IEEE International Conference on Robotics and Automation (ICRA)}, 2018, pp. 7757--7764.

\bibitem{SchwarzLGKPSB:ICRA18}
M.~Schwarz, C.~Lenz, G.~M. Garc{\'{\i}}a, S.~Koo, A.~S. Periyasamy, M.~Schreiber, and S.~Behnke, ``Fast object learning and dual-arm coordination for cluttered stowing, picking, and packing,'' in \emph{{IEEE} International Conference on Robotics and Automation (ICRA)}, 2018, pp. 3347--3354.

\bibitem{armbench}
C.~Mitash, F.~Wang, S.~Lu, V.~Terhuja, T.~Garaas, F.~Polido, and M.~Nambi, ``{ARMBench}: An object-centric benchmark dataset for robotic manipulation,'' in \emph{International Conference on Robotics and Automation (ICRA)}, 2023, pp. 9132--9139.

\bibitem{megapose}
Y.~Labb\'e, L.~Manuelli, A.~Mousavian, S.~Tyree, S.~Birchfield, J.~Tremblay, J.~Carpentier, M.~Aubry, D.~Fox, and J.~Sivic, ``{MegaPose}: {6D} pose estimation of novel objects via render \& compare,'' in \emph{Conference on Robot Learning (CoRL)}, 2022.

\bibitem{maskrcnn}
K.~He, G.~Gkioxari, P.~Doll{\'a}r, and R.~Girshick, ``Mask {R-CNN},'' in \emph{International Conference on Computer Vision (ICCV)}, 2017, pp. 2961--2969.

\bibitem{segment_anything}
A.~Kirillov, E.~Mintun, N.~Ravi, H.~Mao, C.~Rolland, L.~Gustafson, T.~Xiao, S.~Whitehead, A.~C. Berg, W.-Y. Lo \emph{et~al.}, ``Segment anything,'' in \emph{International Conference on Computer Vision (ICCV)}, 2023, pp. 4015--4026.

\bibitem{fast_sam}
X.~Zhao, W.~Ding, Y.~An, Y.~Du, T.~Yu, M.~Li, M.~Tang, and J.~Wang, ``Fast segment anything,'' \emph{arXiv preprint arXiv:2306.12156}, 2023.

\bibitem{template_matching_Hu}
Z.~Hu, R.~Tan, Y.~Zhou, J.~Woon, and C.~Lv, ``Template-based category-agnostic instance detection for robotic manipulation,'' \emph{Robotics and Automation Letters (RA-L)}, vol.~7, no.~4, pp. 12\,451--12\,458, 2022.

\bibitem{robollm}
Z.~Long, G.~Killick, R.~McCreadie, and G.~A. Camarasa, ``{RoboLLM}: Robotic vision tasks grounded on multimodal large language models,'' 2023.

\bibitem{sdmaskrcnn}
M.~Danielczuk, M.~Matl, S.~Gupta, A.~Li, A.~Lee, J.~Mahler, and K.~Goldberg, ``Segmenting unknown {3D} objects from real depth images using {Mask} {R-CNN} trained on synthetic data,'' in \emph{Int. Conf. Robotics and Automation (ICRA)}, 2019.

\bibitem{uois}
C.~Xie, Y.~Xiang, A.~Mousavian, and D.~Fox, ``Unseen object instance segmentation for robotic environments,'' \emph{IEEE Transactions on Robotics (T-RO)}, 2021.

\bibitem{msmformer}
Y.~Lu, Y.~Chen, N.~Ruozzi, and Y.~Xiang, ``Mean shift mask transformer for unseen object instance segmentation,'' \emph{arXiv preprint arXiv:2211.11679}, 2022.

\bibitem{instr}
M.~Durner, W.~Boerdijk, M.~Sundermeyer, W.~Friedl, Z.-C. Márton, and R.~Triebel, ``Unknown object segmentation from stereo images,'' in \emph{International Conference on Intelligent Robots and Systems (IROS)}, 2021, pp. 4823--4830.

\bibitem{beit3}
W.~Wang, H.~Bao, L.~Dong, J.~Bjorck, Z.~Peng, Q.~Liu, K.~Aggarwal, O.~K. Mohammed, S.~Singhal, S.~Som, and F.~Wei, ``Image as a foreign language: {BEiT} pretraining for vision and vision-language tasks,'' in \emph{Conference on Computer Vision and Pattern Recognition (CVPR)}, 2023.

\bibitem{dtoid}
J.-P. Mercier, M.~Garon, P.~Giguère, and J.-F. Lalonde, ``Deep template-based object instance detection,'' in \emph{Winter Conference on Applications of Computer Vision (WACV)}, 2021, pp. 1506--1515.

\bibitem{dounseen}
A.~Gouda and M.~Roidl, ``{DoUnseen}: Tuning-free class-adaptive object detection of unseen objects for robotic grasping,'' 2023.

\bibitem{cnos}
V.~N. Nguyen, T.~Groueix, G.~Ponimatkin, V.~Lepetit, and T.~Hodan, ``{CNOS}: A strong baseline for {CAD-based} novel object segmentation,'' in \emph{International Conference on Computer Vision (CVPR)}, 2023, pp. 2134--2140.

\bibitem{sam6d}
J.~Lin, L.~Liu, D.~Lu, and K.~Jia, ``{SAM-6D}: Segment anything model meets zero-shot 6d object pose estimation,'' \emph{arXiv preprint arXiv:2311.15707}, 2023.

\bibitem{falling_things_dataset}
J.~Tremblay, T.~To, and S.~Birchfield, ``Falling things: A synthetic dataset for {3D} object detection and pose estimation,'' in \emph{Conference on Computer Vision and Pattern Recognition (CVPR) Workshops}, 2018, pp. 2038--2041.

\bibitem{hope_dataset}
S.~Tyree, J.~Tremblay, T.~To, J.~Cheng, T.~Mosier, J.~Smith, and S.~Birchfield, ``{6-DoF} pose estimation of household objects for robotic manipulation: An accessible dataset and benchmark,'' in \emph{International Conference on Intelligent Robots and Systems (IROS)}, 2022.

\bibitem{resnet_rmac}
G.~Tolias, R.~Sicre, and H.~J{\'e}gou, ``Particular object retrieval with integral max-pooling of {CNN} activations,'' \emph{arXiv preprint arXiv:1511.05879}, 2015.

\bibitem{DINO_ViTS}
M.~Caron, H.~Touvron, I.~Misra, H.~J{\'e}gou, J.~Mairal, P.~Bojanowski, and A.~Joulin, ``Emerging properties in self-supervised vision transformers,'' in \emph{International Conference on Computer Vision (ICCV)}, 2021, pp. 9650--9660.

\bibitem{dinov2_registers}
T.~Darcet, M.~Oquab, J.~Mairal, and P.~Bojanowski, ``Vision transformers need registers,'' \emph{arXiv preprint arXiv:2309.16588}, 2023.

\bibitem{learn_from_sam}
M.~E. Bonani, M.~Schwarz, and S.~Behnke, ``Learning from {SAM}: Harnessing a segmentation foundation model for {Sim2Real} domain adaptation through regularization,'' \emph{arXiv preprint arXiv:2309.15562}, 2023.

\bibitem{resnet}
K.~He, X.~Zhang, S.~Ren, and J.~Sun, ``Deep residual learning for image recognition,'' in \emph{Conference on Computer Vision and Pattern Recognition (CVPR)}, 2016, pp. 770--778.

\bibitem{vit}
A.~Dosovitskiy, L.~Beyer, A.~Kolesnikov, D.~Weissenborn, X.~Zhai, T.~Unterthiner, M.~Dehghani, M.~Minderer, G.~Heigold, S.~Gelly, J.~Uszkoreit, and N.~Houlsby, ``An image is worth 16x16 words: Transformers for image recognition at scale,'' in \emph{International Conference on Learning Representations (ICLR)}, 2021.

\bibitem{finetuning_vit}
A.~Kumar, R.~Shen, S.~Bubeck, and S.~Gunasekar, ``How to fine-tune vision models with {SGD},'' \emph{arXiv preprint arXiv:2211.09359}, 2022.

\bibitem{trivial_aug}
S.~G. M{\"u}ller and F.~Hutter, ``{TrivialAugment}: Tuning-free yet state-of-the-art data augmentation,'' in \emph{International Conference on Computer Vision (ICCV)}, 2021, pp. 774--782.

\bibitem{hope_video}
Y.~Lin, J.~Tremblay, S.~Tyree, P.~A. Vela, and S.~Birchfield, ``Multi-view fusion for multi-level robotic scene understanding,'' in \emph{IEEE/RSJ International Conference on Intelligent Robots and Systems (IROS)}, 2021, pp. 6817--6824.

\end{thebibliography}

\end{document}